\documentclass{article}
\usepackage{spconf,amsmath,graphicx}

\usepackage{tikz}
\usepackage{comment}
\usepackage{amsmath,amssymb} 
\usepackage{color}
\usepackage{url}
\usepackage{dsfont}
\usepackage{xcolor}
\usepackage{soul}

\usepackage{adjustbox}

\newcommand\T{^\top}

\usepackage{amssymb}
\newcommand{\loss}{\mathcal{L}}

\newcommand{\R}{\mathds{R}}

\renewcommand{\d}{\partial\,}

\usepackage{amsmath}
\DeclareMathOperator*{\argmax}{arg\,max}

\title{Fine-tune your Classifier: Finding Correlations With Temperature} 

\name{
    \parbox{\linewidth}{\centering
    ~~~~~~~~~~~~~~~~~~~~~~~~Benjamin Chamand$^{1\ast}$ \qquad
    Olivier Risser-Maroix$^{2\ast\dag}$ \newline
    \textit{Camille Kurtz$^2$ \qquad
    Philippe Joly$^1$ \qquad
    Nicolas Loménie$^2$}
   \thanks{$\ast$ Equal contribution.\\ \indent~~~$\dag$ Financed by Smiths Detection}
   }
}

\address{
    \vspace{1mm}$^1$IRIT, Université de Toulouse, CNRS, Toulouse INP, UT3, Toulouse, France\\
    \vspace{0.5mm}$^2$LIPADE, Université de Paris, France\\
    \vspace{0.5mm}
    \normalsize\texttt{benjamin.chamand@irit.fr, orissermaroix@gmail.com}
}

\begin{document}

\maketitle{}

\begin{abstract}
Temperature is a widely used hyperparameter in various tasks involving neural networks, such as classification or metric learning, whose choice can have a direct impact on the model performance.
Most of existing works select its value using hyperparameter optimization methods requiring several runs to find the optimal value.
We propose to analyze the impact of temperature on classification tasks by describing a dataset as a set of statistics computed on representations on which we can build a heuristic giving us a default value of temperature.
We study the correlation between these extracted statistics and the observed optimal temperatures.
This preliminary study on more than a hundred combinations of different datasets and features extractors highlights promising results towards the construction of a general heuristic for temperature.
\end{abstract}

\begin{keywords}
temperature, hyperparameter, heuristic, softmax, cross-entropy
\end{keywords}

\section{Introduction}
\label{sec:intro}

The performance of a machine learning algorithm applied to a computer vision task 
is highly dependent on the choice of its hyperparameters.
Among these, the temperature is a scaling factor often used in a neural network linked to the softmax layers, 
the latter being usually followed by a cross-entropy (CE) like loss function. 
Intuitively, the temperature (in allusion to statistical mechanics) is introduced to choose the level of uniformity of the distribution.
Since most deep classification models involve both softmax layer and CE like loss functions 
for their training, 
determining an optimal temperature for a particular task can then have a broad impact.

For example, this parameter is widely considered in various tasks such as knowledge distillation, classification, text generation, self-supervised and metric learning \cite{hinton2015distilling,zhai2018classification,Caccia2020Language,Hu2017Toward,chen2020simple,guo2017calibration,khosla2020supervised,Wu_2018_ECCV,risser2021learning,abs-2012-13575,MaYLNH17} 
Traditionally, in most of these domains and in the underlying applications, the temperature is determined empirically, with a value that can be constant (typically from a grid search) or evolve dynamically over iterations, 
in the same vein as the learning rate parameter. 
Nevertheless, such strategies for determining a \textit{good} temperature may be suboptimal or computationally too cumbersome. 
Surprisingly, there are very few studies proposing strategies for determining an optimal temperature.
In this paper, we focus on the particular problem that, given a classification task, we need to find a correlation between an optimal value for the temperature and statistics describing the dataset such as complexity, dimension, number of classes, etc.

\section{Related Works}
\label{sec:SOTA}

The temperature hyperparameter is typically employed in the softmax layer to control the uniformity of the distribution.
Although the use of a good temperature has shown its impact in many computer vision tasks, the existing strategies to define such a temperature parameter are quite limited.

The first way to proceed is to consider a constant temperature throughout the training.
The choice can be done empirically, as in \cite{hinton2015distilling, Wu_2018_ECCV, risser2021learning}.
It can also be considered as a fixed hyperparameter to be optimized via a grid search in a field of possible values, but this implies significant computational requirements 
 and leads to different hyperparameters for each dataset and architecture.
A simple heuristic can also allow to fix the parameter as proposed in the Transformers \cite{vaswani2017attention} with $\sqrt{d}$, $d$ being the dimension of the queries and the key vectors.

Other strategies rely on dynamic temperature adjustment during learning iterations. In this case, the elements of the temperature can evolve at each epoch using a scheduler \cite{Hu2017Toward}, in the manner of the learning rate to refine the network.
In \cite{zhang2018heated}, the authors also showed that a batch normalization rescaled by $\sqrt{d}$, with $d$ the number of dimensions of embeddings, worked slightly better than a simple $L2$ normalization, and can also lead to more embedding vectors.
Dynamic adjustment of temperature can also be done by learning it as a standard parameter \cite{radford2021learning,Ranjan2017L2constrainedSL}. This usually requires additional steps like clipping or adding \textit{exp} to avoid negative values. 
Furthermore, the learned temperature strongly depends on the learning rate hyperparameter. 

An alternative approach is to determine the temperature value analytically.
The authors of \cite{HE201880} propose both an evaluation function designed to measure the effectiveness of a temperature parameter and an iterative updating rule to determine the optimal temperature value.
However, their work suffers from two drawbacks: (1) Authors introduce a novel hyperparameter $\lambda$ in the temperature formulation, $\lambda$ being an improvement factor affecting the number of iterations and the selection of the optimal temperature; (2) It was designed for the \textit{$\mathcal{D}$-armed bandit} problem in reinforcement learning and only tested on synthetic data.

Another work by \cite{liu2017rethinking} proposes a theoretical lower bound formulated as a function of the loss value and the number of classes with a loss smaller than $\epsilon$, $\epsilon$ supposed to be around $10e-4$.
Interestingly, unlike \cite{vaswani2017attention,zhang2018heated}, their solution does not derive any benefit from or rely on any information on the embeddings dimensions.
However, since temperature determination was not the main part of their contribution, no benchmark was made to compare the proposed theoretical lower bound with other temperature values.
Finally, the assumption on such a low loss value does not correspond to real cases at the beginning of the learning.

While some of the previously mentioned heuristics are based on feature dimensions, others use the number of classes or class separability measures.
None of them have been designed specifically for use in a classification task or have been evaluated on this particular hyperparameter to demonstrate the effectiveness of the proposed heuristic.
Other heuristics could be derived from other criteria reflecting information such as the difficulty of the dataset to be classified.
For example, \cite{lorena2019complex} proposes to estimate the difficulty of classifying datasets from six classes of measures based on information such as feature-based, neighborhood or dimensionality measures. 
However, most of proposed measures have a complexity at least equal to $O(n^2)$, with $n$ the number of points in the dataset, making these measures difficult to scale up to larger datasets.
Similarly, we seek to describe each dataset by a set of statistics computable in a reasonable amount of time. 
We then propose to determine which variables / statistics are really correlated with the best empirical temperature, in order to propose a simple heuristic based on these 
dataset statistics.

\section{Methodology}
\label{sec:methodology}

\subsection{Rescaling Cross-Entropy with temperature scaling}
\label{subsec:CE_correc}

Inspired by the formulation of \cite{zhang2018heated}, we start from the same basic observation as they do.
We define a set of $N$ samples labeled $\{(x_1, y_1), \dots, (x_N, y_N)\}$, where $x_i \in \R^{d}$ is the representation (embedding) of the $i$-th sample, $d$ being the dimensionality of $x_i$, and $y_i \in \{1, \dots, C\}$ is the category label of the sample $x_i$, $C$ being the total number of categories.
Let us consider $W = [w_1, \dots, w_C ]$ where $w_j \in \R^{d}$ is the weights associated with the class $C_j$, we define 
$z_i = x_i W$ with $i \in \{1, \dots, N\}$. 
In our case, we focus on the learning of the weights $W$. 
In the same vein as \cite{zhai2018classification,khosla2020supervised,Wu_2018_ECCV,liu2017rethinking}, we removed the bias term, and we consider the inputs $x$ and weights $W$ $L_2$ normalized. We optimize the cosine similarity since this choice is both popular in classification and metric learning.

The probability that a sample $x$ belongs to the category $c \in \{1, \dots, C\}$ can be predicted by the softmax function as:
\begin{equation}
    p(c | x, \alpha) = \frac{\exp(\alpha z_c)}{\sum_{j=1}^{C} \exp(\alpha z_j)}
\end{equation}
To simplify the notation, we note $\alpha = 1 / T$ as the inverse of the temperature $T$ to choose the level of uniformity of the softmax output distribution. 


Assuming that the ground truth distribution of the training sample is $q(c|x)$, generally encoded in a one-hot vector (which equals 1 if $c = y$ and 0 otherwise), the cross-entropy loss with respect to $x$ is defined as:
\begin{equation}
    \loss(x, \alpha) = - \sum_{c=1}^{C} \log(p(c | x, \alpha)) q(c | x)
\end{equation}
and the gradient with respect to the weight $w_c$ is:
\begin{equation}
    \label{eq:gradient_wrt_w}
    \frac{\d \loss}{\d w_c} = \alpha  (p(c | x, \alpha) - q(c | x)) x\T 
\end{equation}

From  Eq.~\ref{eq:gradient_wrt_w}, we can observe that the temperature has two effects in the gradients.
The first is, as mentioned earlier, to control the probability distribution $p(c|x,\alpha) \in [0, 1]$.
The second effect is simply to multiply gradients by the value of the temperature $\alpha$.
As $\alpha$ often lies (empirically) in $[1, 250]$, this last effect is harmful during learning since it rescales the learning rate with this value; this can lead to divergence.
To cancel this effect and study only the impact of the choice of distribution during training, we propose to normalize the cross-entropy loss function by a constant of value equal to $\alpha$.

\subsection{Finding correlations}

As previously mentioned and illustrated, the temperature has a strong impact on the final accuracy but poses different difficulties in finding the optimal value.
We then look for a heuristic $h$, a universal rule, to select a temperature $\alpha$ close to its optimal value, denoted $\alpha^*$, that is general across datasets and representations. 
We need to represent each dataset of embeddings $e \in \mathcal{E}$ in a common space $\mathcal{S}$ by using some statistical features $s \in \R^{m}$ computed over $\mathcal{E}$ with $m$ the number of statistical features.
Our goal is to find the best correlation with the observed optimal temperature in order to design a heuristic for temperature. 

To ensure our heuristic will be sufficiently general and not just specialized for particular cases such as small number of classes or large embedding sizes, we need to cover as many as possible different cases.
To this end, we construct a list of datasets with different numbers of classes and a list of features extractors with different feature sizes / discrimination powers.
From each pair (dataset, feature extractor) we build a dataset of embeddings $e_i$ divided in training and validation sets $e_i^{train}, e_i^{val}$.
For each set of embeddings $e_i$, we compute a description by extracting the statistical features $s$ and empirically find the corresponding optimal temperature $\hat{\alpha}^*_i = \argmax_{\alpha, W_\alpha} Accuracy(e_i^{val})$ for $\alpha$ selected from a given set of possible temperatures 
and $W_\alpha$ the weights learned on $e_i^{train}$ with a given temperature $\alpha$.
Thus, to each pair (dataset, feature extractor) is associated the pair (embedding dataset statistical features, optimal empiric temperature) noted: $(\mathcal{S}_i, \hat{\alpha}^*_i)$.

In order to find our heuristic $h(\cdot)$ several options are possible.
The simplest one would be to consider an affine function on the form $a \cdot s_j+b$, with $s_j$ the most correlated variable in $s$ to our optimal temperature.
If a 
strong enough
correlation exists, this would be the simplest heuristic possible.
However, more than one variable may be needed to find this correlation.
We therefore investigate the strength of the correlation between each statistic independently and a linear combination of our statistical features  to the optimal temperature.

\section{Experimental Study}
\label{sec:experiments}

\subsection{Datasets, feature extractors selection}

\begin{figure}
    \centering
    \includegraphics[width=.33\textwidth]{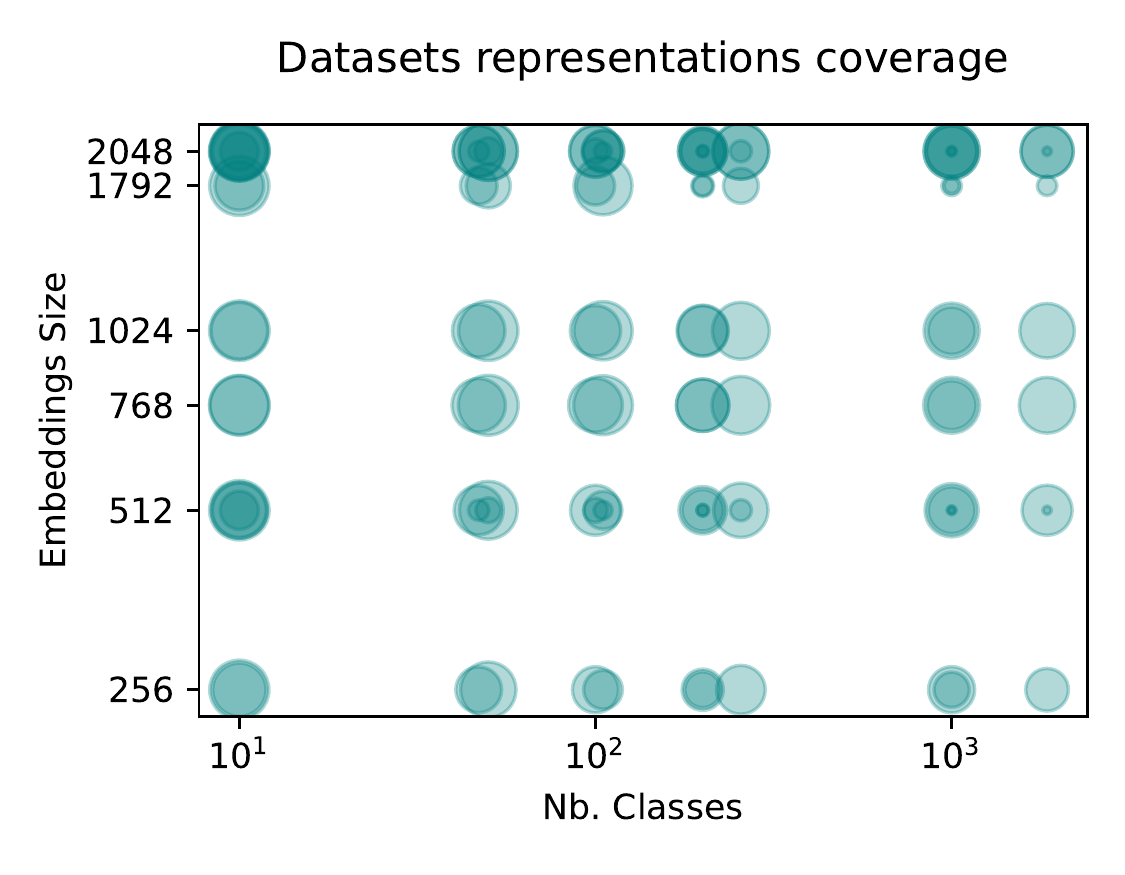}
    \caption{Coverage between all combinations of feature extractors and datasets. The size of the circles represents the best accuracy achieved for each dataset.}
    \label{fig:datasets_coverage}
\end{figure}

In order to find a generalizable heuristic covering a wide range of cases for a classification task, we selected 12 datasets and 9 feature extractors (Fig.~\ref{fig:datasets_coverage}).
The number of classes ranges from 10 to 1854 while the dimensionality of the features ranges from 256 to 2048. 
The selected datasets are MNIST
, CIFAR10
, DTD
, PhotoArt
, CIFAR100
, 105-PinterestFaces
, CUB200
, ImageNet-R
, Caltech256
, FSS1000
, ImageNetMini
, THINGS
, containing respectively 10, 10, 47, 50, 100, 105, 200, 200, 256, 1000, 1000, 1854 classes.
Regarding the feature extractors, different architectures have been selected with different pre-trainings, in order to cover a large number of dimensions while decorrelating this aspect from the network performance.
For example, FaceNet is expected to perform poorly on CIFAR datasets since it is learned on a face recognition task while a ResNet18 pretrained on ImageNet is expected to perform better on natural images while being weaker on 105-PinterestFaces.
The feature extractors used are: AlexNet \cite{krizhevsky2012imagenet} and ResNet-\{18, 50, 101\} \cite{he2016deep} pre-trained on ImageNet, ResNet-\{34, 152\} \cite{he2016deep} randomly initialized, FaceNet \cite{schroff2015facenet} pre-trained on VGGFaces2 
and CLIP-\{RN50, ViT32\_b\} \cite{radford2021learning} pre-trained on millions of image-text pairs.
The embedding dimensions are respectively: 256, 512, 2048, 2048, 512, 2048, 1792, 1024, 768.

\begin{figure}[!t]
    \centering
    \includegraphics[width=.44\textwidth]{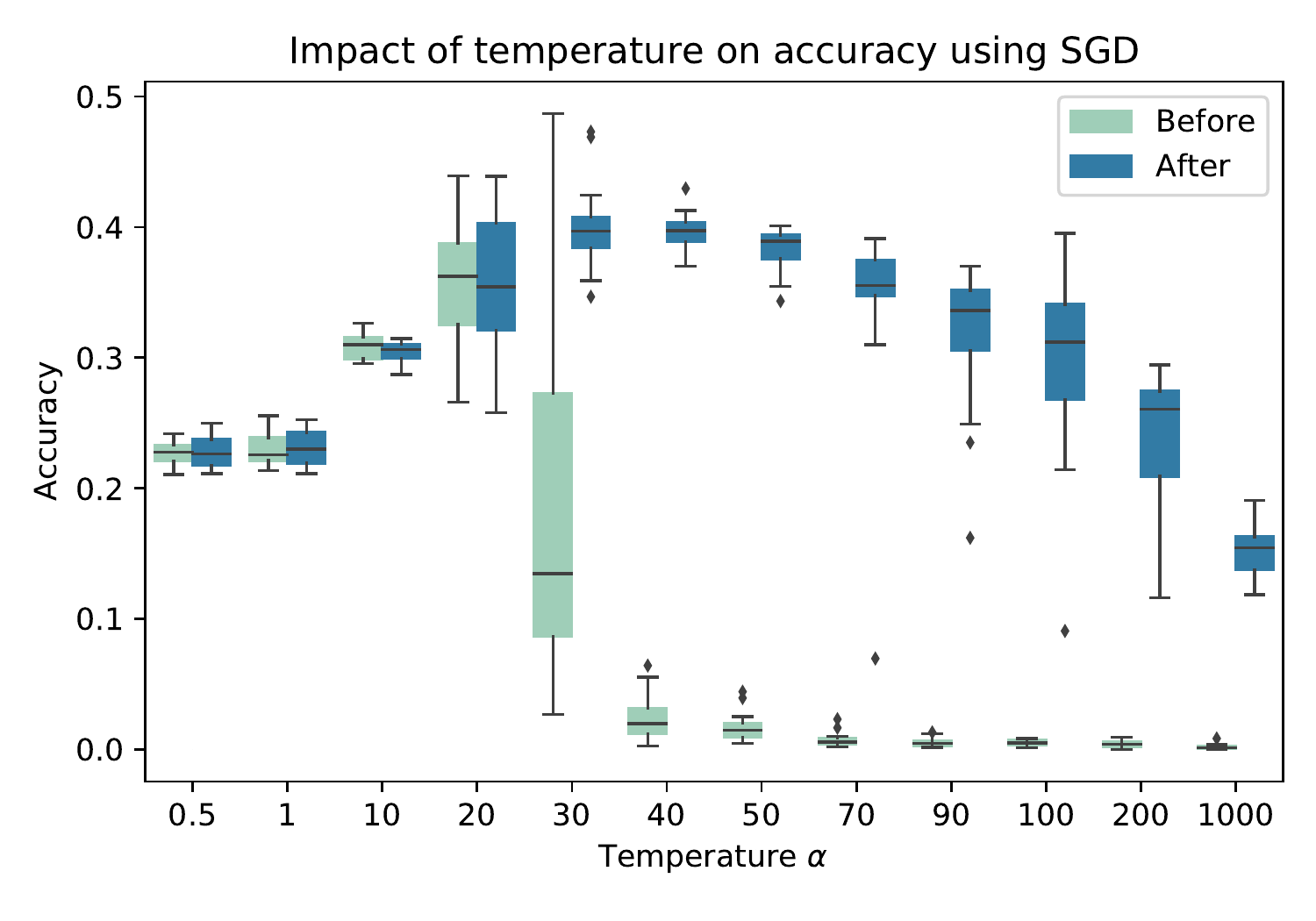}
    \vspace{-1mm}
    \caption{Impact of temperature on the accuracy of the dataset THINGS \cite{Hebart2019THINGS}. For each temperature, we learned the classification layer and performed 20 learning for 100 epochs.}
    \label{fig:temperature_CE_normalized_THINGS}
\end{figure}

\subsection{Statistical features selection}

In order to find the hidden relationship between a given dataset and the associated optimal temperature we need to describe each dataset by a feature vector $s$ in a common space $\mathcal{S}$.
Since, as we have seen previously, very different heuristics are proposed to set up the temperature, we selected various features $s_i$:
the dimensionality of embeddings $e$ (\textit{dim}), 
the number of output classes (\textit{n\_classes}), 
the mean value of all embeddings values (\textit{mean}), 
the variance of all embedding values (\textit{var}), 
the trace of the average matrix of all intra-class covariance matrices (\textit{sb\_trace}), 
the trace of the average of all inter-classes covariances matrices (\textit{sw\_trace}), 
the mean squared error (MSE) between the features correlation matrix and the identity (\textit{feats\_corr}), 
the mean cosine similarity between each dimensions pair (\textit{feats\_cos\_sim}),
the number of samples in the training set (\textit{n\_samples}), 
the average number of samples per class (\textit{avg\_samp\_class}) and the percentage of dimensions to be retained for a given explained variance (as in PCA) of 50, 75, 80, 90, 95, 99 (\textit{pca\_\%}), the average of all embedding values (\textit{train\_mean}) and the standard deviation (\textit{train\_std}), the average kurtosis computed on each dimension (\textit{avg\_kurtosis}), and the average Shapiro-Wilk value testing the normality of each dimension (\textit{avg\_normality}).
Three other popular metrics used in clustering are used such as the Silhouette (\textit{silhouette}), Calinski Harabasz (\textit{calinski\_harabasz})   and the David Bouldin (\textit{david\_bouldin}) score using the true labels as cluster prediction to obtain measures of the quality of the representation.

\subsection{Empiric study of correlations}

\begin{figure}
    \centering
    \includegraphics[width=.48\textwidth]{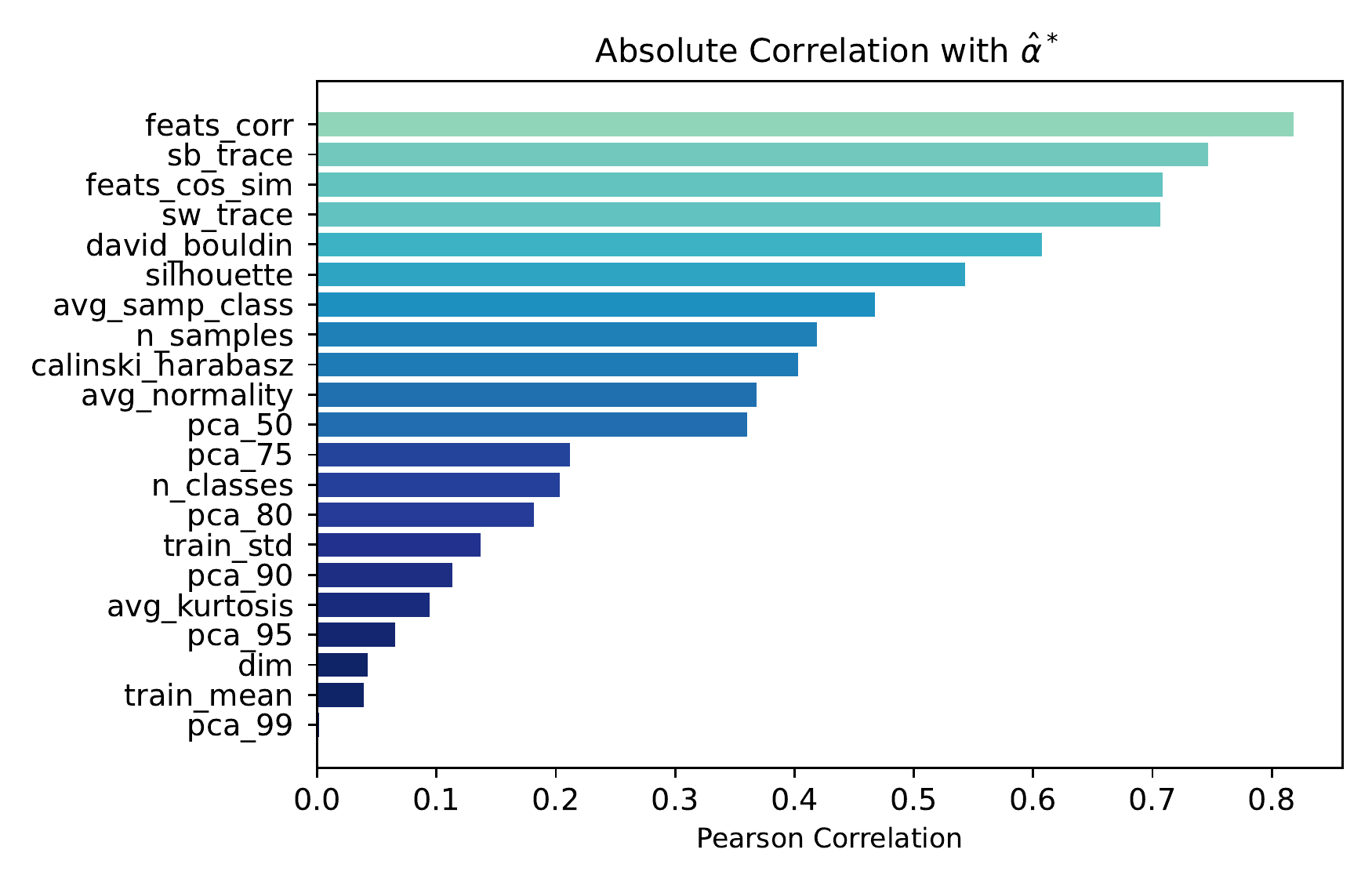}
    \caption{Absolute value of Pearson correlation between each dataset statistic and the optimal empirical temperature.}
    \label{fig:abs_corr_temp}
\end{figure}

Once we have extracted the embeddings from various datasets and feature extractors, we need to find the best temperature usable for each case.
To do this, we split each dataset of embeddings into training and testing sets 
and trained the model during 1000 epochs with a batch of size 2048 for a given temperature.
We used the default temperature of 1 and temperatures ranging from 5 to 250 with a step of 5: $\alpha \in \{1, 5, \dots, 245, 250\}$.
By tracking the accuracy on the test set, we are able to observe the best achievable accuracy for each temperature.
We used the rescaled CE loss presented in Sec.~\ref{subsec:CE_correc} which allows strong improvements in accuracy over high temperatures using the SGD optimizer as shown in Fig.~\ref{fig:temperature_CE_normalized_THINGS}. The latter allows us to observe experimentally the advantage of isolating the peaking distribution effect.
However, we found that this had no impact during training when using a smarter optimizer like Adam \cite{kingma2014adam}.

\begin{figure}
    \centering
    \includegraphics[width=.42\textwidth]{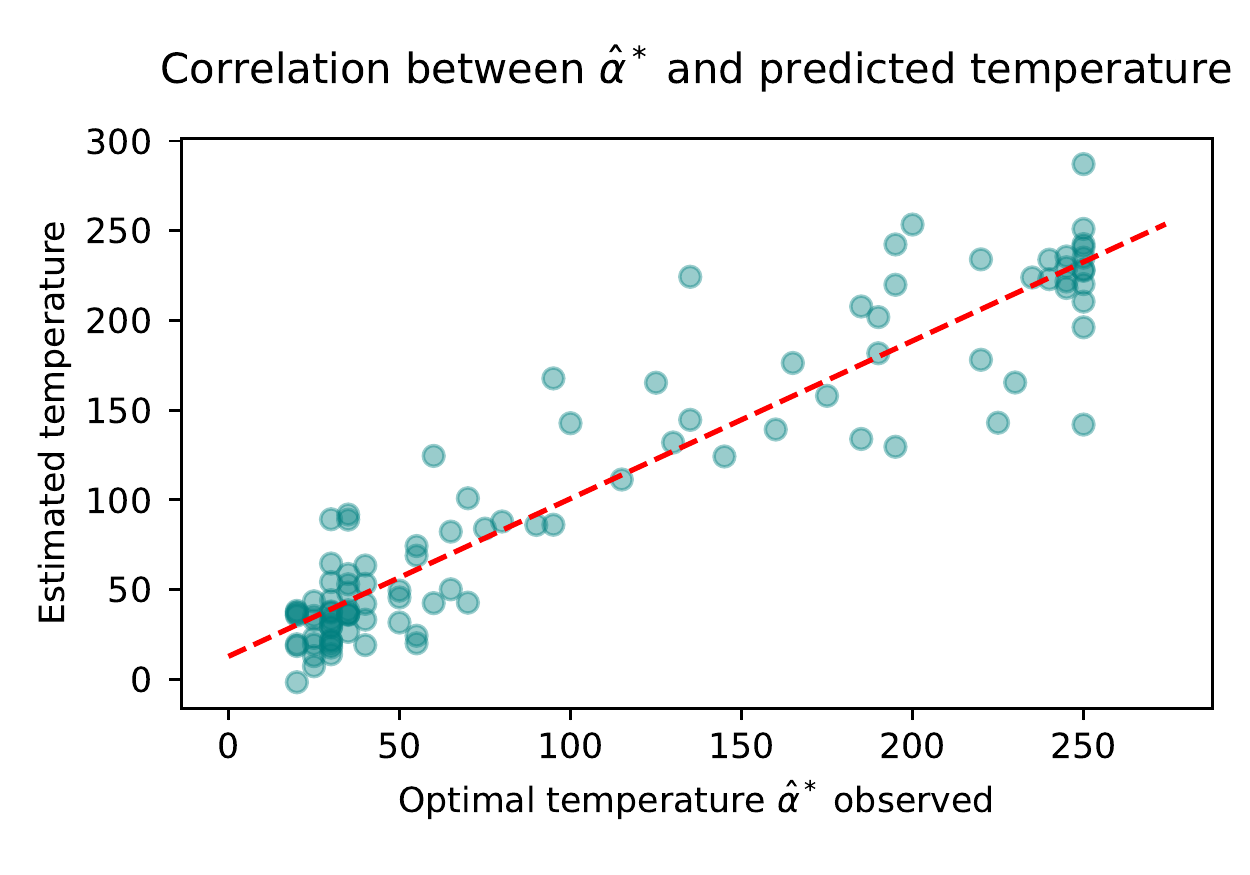}
    \caption{Correlation between observed optimal temperature versus predicted temperature.} 
    \label{fig:correlation_true_predicted_temp}
    \vspace{-1mm}
\end{figure}

In order to find a heuristic for setting a default temperature, we need to find strong correlations from the pairs of optimal empiric temperatures and datasets statistics.
Fig.~\ref{fig:abs_corr_temp} shows the absolute correlation between each statistical value and the temperature.
We found that the most interesting variable was the measure of correlation between embedding features. 
To increase the correlation, we propose to learn a linear regression from our statistics and the optimal temperature, using a cross-validation strategy. The latter omits all sets of embeddings of a dataset (e.g. MNIST) during the learning phase in order to use them in the validation of the found linear combination.
After that, we repeated this procedure on a subset of the most correlated statistics.
The scores are shown in Tab.~\ref{tab:cross_val_correlation}.
Finally, we fitted a linear regression on all points whose correlation between our predicted temperature and the empirical optimal temperature is shown in Fig.~\ref{fig:correlation_true_predicted_temp}
Obtained results are promising
and statistically significant with a Pearson's correlation of 0.9563 and a $p$-value of $0.014 < 0.05$.

\begin{table}
\centering
\begin{adjustbox}{width=1\columnwidth,center}
   \begin{tabular}{ | l || c | c | r | }
     \hline
     Method & Med. Corr. & Avg. Corr. ($\pm$ std) & $p$-value \\
     \hline
     \hline
     Best Stat & 0.9262 & 0.851 ($\pm$ 0.159) & 0.031 \\ \hline
     4-Best Stats & 0.9265 & 0.865 ($\pm$ 0.131) & 0.018 \\ \hline
     All Stats & \textbf{0.9563} & 0.884 ($\pm$ 0.124) & 0.014 \\
     \hline
   \end{tabular}
   \end{adjustbox} \\
   \caption{Observed correlations between the most correlated variable, a linear combination of 4 variables and a linear combination of all variables with the optimal empiric temperature.}
  \label{tab:cross_val_correlation}
\end{table}

\section{Conclusion}
\label{sec:conclusion}

In this paper, we have shown the importance of the temperature hyperparameter for finetuning a linear classifier on learned representation.
We showed that cross-entropy loss can suffer from high temperature if not properly re-scaled.
After re-scaling the cross-entropy, we proposed to study the correlations between the optimal empirical temperature observed on many datasets, over a wide range of classes and dimensions, and the statistics computed on the representations of the dataset.
In this way, we revealed that some heuristics (such as the dimensionality of embeddings) had little correlation with the optimal temperature while a measure of correlation between features showed strong correlations.
We found that appropriate selection and combination of statistics could improve the correlation with the best temperature.
We suggest enhancing this pipeline in subsequent work \cite{risser2022can} and applying it to other issues, such as identifying the elements of representation learning that will result in high accuracy by predicting it with symbolic regression.

\bibliographystyle{IEEEbib}
\bibliography{strings,refs}

\end{document}